\documentclass[12pt]{article}

\usepackage[margin=1in]{geometry}
\usepackage{setspace}
\usepackage[utf8]{inputenc}
\usepackage[T1]{fontenc}
\usepackage{mathpazo}  

\usepackage{amsmath,amssymb}
\usepackage{graphicx}
\usepackage{booktabs}
\usepackage{array}
\usepackage{multirow}
\usepackage{xcolor}
\usepackage{tikz}
\usetikzlibrary{arrows.meta,positioning,shapes.geometric}
\usepackage{natbib}
\usepackage[colorlinks=true,linkcolor=blue!60!black,citecolor=blue!60!black,urlcolor=blue!60!black]{hyperref}
\usepackage{url}

\title{\textbf{Vibe Researching as Wolf Coming: \\[0.3em] Can AI Agents with Skills Replace or Augment Social Scientists?}}

\author{Yongjun Zhang\thanks{Department of Sociology and Institute for Advanced Computational Science, Stony Brook University. Email: yongjun.zhang@stonybrook.edu. The author used scholar-skill, a Claude Code plugin, as an AI research assistant during the preparation of this manuscript. All content was reviewed, revised, and verified by the author, who takes full intellectual responsibility. The author declares no competing interests.}}

\date{March 2026}

\begin{document}

\maketitle

\begin{abstract}
\noindent AI agents---systems that execute multi-step reasoning workflows with persistent state, tool access, and specialist skills---represent a qualitative shift from prior automation technologies in social science. Unlike chatbots that respond to isolated queries, AI agents can now read files, run code, query databases, search the web, and invoke domain-specific skills to execute entire research pipelines autonomously. This paper introduces the concept of \textit{vibe researching}---the AI-era parallel to ``vibe coding'' \citep{karpathy2025vibe}---and uses scholar-skill, a 26-skill plugin for Claude Code covering the full research pipeline from idea to submission across 18 orchestrated phases with 53 quality gates, as an illustrative case. I develop a cognitive task framework that classifies research activities along two dimensions---codifiability and tacit knowledge requirement---to identify a delegation boundary that is cognitive, not sequential: it cuts through every stage of the research pipeline, not between stages. I argue that AI agents excel at speed, coverage, and methodological scaffolding but struggle with theoretical originality and tacit field knowledge. The paper concludes with an analysis of three implications for the profession---augmentation with fragile conditions, stratification risk, and a pedagogical crisis---and proposes five principles for responsible vibe researching.

\end{abstract}

\newpage

\section{Introduction}

When was the last time you did something in your research that only \textit{you} could have done? Not ran the code. Not wrote the literature review. Not formatted the citations. Something that required your specific judgment---your field knowledge, your theoretical imagination, your sense of what matters.

The question is not rhetorical. AI systems have demonstrated the capacity to automate a wide range of knowledge work \citep{eloundou2024gpts}, from code generation (2022) to coherent text production (2023) to tool-using multi-step workflows (2024) \citep{park2023generative, wu2023autogen}. In early 2025, Andrej Karpathy coined the term ``vibe coding'' to describe a programming style in which the user describes what they want, the AI writes the code, and the user accepts it without reading the diffs \citep{karpathy2025vibe}. The concept has since spawned a parallel in academia: ``vibe researching,'' where scholars describe what they want to study and AI agents handle the execution \citep{pachocki2025vibecodingresearching}. Tools like SciSciGPT, an open-source prototype AI collaborator, reportedly complete many research tasks substantially faster than human researchers \citep{shao2025sciscigpt}. This paper names and analyzes this phenomenon in social science: \textit{vibe researching}. The title's ``wolf coming'' alludes to the Chinese fable of the boy who cried wolf---but in this version, the wolf is real. You describe your research question. The AI runs the literature review, designs the study, analyzes the data, drafts the paper, and simulates the reviewers.

The core question that structures this paper is: when AI agents can execute the research pipeline, what is the researcher's distinctive contribution? Is the researcher still the \textit{author}? The \textit{designer}? Or are they becoming something more like a \textit{curator}---selecting among options the AI generates?

I address this question by developing a cognitive task framework and applying it to a concrete case: scholar-skill, a 26-skill plugin for Claude Code that covers the entire social science research workflow, from initial puzzle to journal submission. The system is calibrated for publication in twenty-two journals spanning sociology, demography, political science, computational social science, and interdisciplinary science, including the \textit{American Sociological Review}, the \textit{American Journal of Sociology}, \textit{Demography}, \textit{Social Forces}, \textit{Science Advances}, \textit{Nature Human Behaviour}, \textit{Nature Computational Science}, \textit{Language in Society}, \textit{Gender \& Society}, the \textit{American Political Science Review}, \textit{Journal of Marriage and Family}, \textit{Population and Development Review}, \textit{Sociological Methods \& Research}, \textit{Poetics}, \textit{PNAS}, \textit{Social Science Research}, \textit{Journal of Mixed Methods Research}, \textit{Sociological Theory}, the \textit{Annual Review of Sociology}, \textit{Journal of Sociolinguistics}, \textit{Journal of Quantitative Linguistics}, and \textit{Sociological Forum}.

The argument proceeds in five steps. First, I situate the current moment within four historical waves of research automation, showing that the present wave is qualitatively different because it automates \textit{reasoning across stages}, not merely execution within a single stage. Second, I describe the scholar-skill system as an illustrative case of agentic AI in social science. Third, I develop a two-dimensional framework---codifiability and tacit knowledge---that identifies a delegation boundary cutting through every stage of the research pipeline. Fourth, I assess what AI agents can and cannot do, distinguishing speed and methodological scaffolding (where AI excels) from theoretical originality and tacit field knowledge (where it struggles). Fifth, I draw out three implications: the augmentation thesis and its fragile conditions, the stratification risk of an ``AI productivity premium,'' and a pedagogical crisis in graduate training.

This paper contributes to a growing literature on AI and social science \citep{bail2024generative, ziems2024llm, zhang2023generative} by offering not an empirical test but a conceptual framework grounded in an operational system. The framework's value lies in specifying \textit{where} the human--AI boundary falls---not between pipeline stages, but within them---and in naming the conditions under which augmentation succeeds or fails.

\section{Background: From Automation to Agentic AI}

\subsection{Four Waves of Research Automation}

Social science research has been progressively automated across four waves, each targeting a different type of cognitive work (Table~\ref{tab:waves}). This periodization builds on task-based frameworks in labor economics that distinguish routine from non-routine work \citep{autor2015why, acemoglu2019automation}.

\begin{table}[h]
\centering
\caption{Four Waves of Research Automation in the Social Sciences}
\label{tab:waves}
\small
\begin{tabular}{llll}
\toprule
\textbf{Wave} & \textbf{Period} & \textbf{What was automated} & \textbf{What remained human} \\
\midrule
1 & 1970s--1990s & Computation (SPSS, Stata) & Reasoning, interpretation \\
2 & 2000s & Data collection (web, APIs) & Design, interpretation \\
3 & 2010s & Text analysis (NLP, ML) & Theory, framing \\
4 & 2024+ & Multi-step reasoning & \textit{Open question} \\
\bottomrule
\end{tabular}
\end{table}

The first wave (1970s--1990s) automated computation: statistical software replaced manual calculation but left reasoning, interpretation, and design entirely human \citep{grimmer2022text}. The second wave (2000s) automated data collection: web scraping, online surveys, and administrative data linkages expanded the volume of accessible data without changing how researchers designed studies or interpreted results. The third wave (2010s) automated text analysis and pattern recognition: natural language processing, machine learning classifiers, topic models, and embedding-based methods enabled the computational analysis of text at scale \citep{lazer2009computational, grimmer2022text, nelson2020computational, rodriguez2022word}. In each wave, a rule-based execution task was handed off, but the reasoning---the interpretation, the design, the theory---remained human.

The fourth wave, beginning around 2024, is different. For the first time, the thing being automated is \textit{reasoning itself}---multi-step inference, argument construction, methodological judgment. Large language models (LLMs) now generate not only text but also research designs, causal diagrams, identification strategies, and manuscript sections calibrated to specific journal norms \citep{bail2024generative}. If the personal computer was a ``bicycle for the mind,'' AI agents are ``aeroplanes for the mind''---they amplify cognitive reach far beyond what earlier tools offered, but with proportionally greater risks when errors occur \citep{wang2026aeroplanes}. The discourse around this transformation has polarized: accelerationists argue that academia must urgently adapt or face irrelevance \citep{kustov2026wakeup}, while cautious voices warn of deskilling and erosion of scientific norms \citep{hosseini2026benefits}. The question of what remains irreducibly human after Wave 4 is genuinely open.

\subsection{From Chatbot to AI Agent}

The distinction between a chatbot and an AI agent is architectural, not merely rhetorical. A chatbot processes a single query and returns a response. It is stateless: it has no memory across exchanges, no file access, no code execution capability, and no tool use.

An AI agent, by contrast, is multi-step, persistent, and tool-using \citep{park2023generative, wu2023autogen}. In the context of social science research, this means the agent can: read and write files to disk; execute R, Python, or Stata code via a shell; query databases (Zotero, NHANES, IPUMS); search the web and fetch papers; spawn parallel sub-agents for independent tasks; and invoke specialist skills with domain-specific knowledge bases. This is a qualitatively different architecture that enables a qualitatively different range of tasks \citep{shinn2024reflexion}.

The inflection from assistant to agent is the critical threshold. Prior AI tools helped researchers \textit{within} a stage---write better code, classify text, check grammar. AI agents can now operate \textit{across} stages, maintaining state from literature review through analysis to manuscript drafting. This cross-stage persistence is what makes ``vibe researching'' possible and what distinguishes the current moment from the chatbot era.

\subsection{Related Work}

Several recent contributions address the intersection of AI and scientific research. \citet{wang2023scientific} provide a comprehensive review of AI across scientific disciplines, mapping capabilities from molecular simulation to materials discovery. \citet{lu2024aiscientist} demonstrate an AI system capable of generating research ideas, writing code, running experiments, and producing full scientific papers autonomously, though with significant quality limitations. \citet{boiko2023autonomous} show that LLM-driven agents can autonomously plan and execute chemistry experiments using robotic laboratories. In the social sciences specifically, \citet{bail2024generative} argues that generative AI can augment social science across data collection, analysis, and theory development, but warns of risks including speculative theorizing untethered from empirical grounding and reduced incentives for deep methodological training. \citet{ziems2024llm} provide a systematic evaluation of LLM capabilities for computational social science tasks, finding strong performance on structured tasks but limitations on those requiring deep domain knowledge. \citet{argyle2023out} demonstrate that LLMs can simulate human survey responses with surprising fidelity, raising both methodological opportunities and validity concerns.

This paper extends this literature by offering a detailed case study of an operational system---scholar-skill---and developing a cognitive task framework that specifies the human--AI delegation boundary with greater precision than prior work. Rather than asking ``can AI do social science?'' in general terms, I ask: which specific cognitive tasks within each pipeline stage can be delegated, and which cannot?

\section{The Scholar-Skill System: An Illustrative Case}

\subsection{Architecture and Scope}

Scholar-skill is a plugin for Claude Code---Anthropic's command-line interface for the Claude language model---that organizes 26 specialist AI skills into a complete research pipeline. The system is installed locally, runs on the researcher's laptop, and accesses local files, databases, and code environments. Table~\ref{tab:skills} summarizes the thirteen stage-groups and their constituent skills.

\begin{table}[h]
\centering
\caption{Scholar-Skill: 26 Skills Organized by Research Stage}
\label{tab:skills}
\small
\begin{tabular}{l>{\raggedright\arraybackslash}p{0.7\textwidth}}
\toprule
\textbf{Stage} & \textbf{Skills} \\
\midrule
Formulation & \texttt{scholar-idea}, \texttt{scholar-lit-review}, \texttt{scholar-lit-review-hypothesis}, \texttt{scholar-hypothesis} \\
Design & \texttt{scholar-design}, \texttt{scholar-causal} \\
Data & \texttt{scholar-data}, \texttt{scholar-eda} \\
Analysis & \texttt{scholar-analyze}, \texttt{scholar-compute}, \texttt{scholar-ling}, \texttt{scholar-qual} \\
Writing & \texttt{scholar-write}, \texttt{scholar-citation} \\
Ethics/Safety & \texttt{scholar-ethics}, \texttt{scholar-safety} \\
Submission & \texttt{scholar-journal}, \texttt{scholar-respond}, \texttt{scholar-open} \\
Replication & \texttt{scholar-replication} \\
Teaching & \texttt{scholar-teach} \\
Collaboration & \texttt{scholar-collaborate} \\
Quality Assurance & \texttt{scholar-auto-improve} \\
Extensions & \texttt{scholar-grant}, \texttt{scholar-presentation}, \texttt{scholar-full-paper} \\
\bottomrule
\end{tabular}
\end{table}

Each skill is a self-contained workflow with its own knowledge base: reference documents, code templates, journal-specific norms, and quality checklists. The orchestrator---\texttt{scholar-full-paper}---coordinates all 26 skills across 18 phases (from a pre-flight data safety gate through idea formalization, literature review, design, data, EDA, analysis, specialized computational or sociolinguistic branches, manuscript drafting, citation harmonization, submission packaging, ethics compliance, peer review simulation, final assembly, optional grant extension, post-rejection resubmission, academic presentation generation, and automatic quality audit) with 53 quality gate items and five hard stops (at data safety, literature and theory verification, pre-draft gate, citation verification, and ethics compliance) that prevent the pipeline from advancing without meeting minimum standards.

\subsection{Key Technical Capabilities}

Several technical capabilities merit description because they illustrate the distance between current AI agents and prior tools.

\textbf{Idea formalization.} \texttt{Scholar-idea} converts a broad research intuition into a formal, testable research question through a multi-step workflow: puzzle clarification, candidate angle generation, quick literature scan against actual databases, research question formalization with population--context--mechanism structure, variable and mechanism mapping, hypothesis derivation, data source inventory, and a five-agent multi-agent evaluation panel (theorist, methodologist, domain expert, journal editor, and devil's advocate) that stress-tests the proposed research questions before the researcher commits to a design. A synthesizer aggregates the panel's ratings into a consensus scorecard, and the questions are refined before final selection.

\textbf{Literature synthesis at scale.} \texttt{Scholar-lit-review-hypothesis} integrates literature review and theory development into a single coherent workflow across five internal phases: literature search (with an incremental search log persisted to disk after every query to prevent data loss from context compaction), synthesis into a six-bin literature map (established findings, contested findings, null findings, mechanisms, methodological landscape, and a precise gap statement naming the closest prior paper), framework selection from 25+ candidates evaluated against the identified gap, hypothesis derivation with explicit mechanism specification (X $\rightarrow$ M $\rightarrow$ Y under conditions C), and production of publication-ready integrated prose calibrated to the target journal's citation density and length conventions. The search phase queries the researcher's local reference library (Zotero, Mendeley, BibTeX, or EndNote---auto-detected), runs web searches across five query types (core topic, mechanism, methods/data, recent frontier, contested terrain), and checks Annual Reviews for canonical works.

\textbf{Causal identification.} \texttt{Scholar-causal} constructs causal DAGs with adjustment sets, selects from thirteen identification strategies (OLS, difference-in-differences, regression discontinuity, instrumental variables, panel fixed effects, matching, synthetic control, causal mediation, DML/causal forests, bunching estimation, shift-share/Bartik IV, and distributional/quantile methods), generates working code in both R and Stata, produces assumption statements and diagnostic tests, and writes the identification argument for the Methods section.

\textbf{Statistical analysis with outcome-type dispatch.} \texttt{Scholar-analyze} implements an outcome-type dispatch system that routes to specialized estimation pipelines based on eleven outcome types (continuous, binary, ordinal, multinomial, count, zero-inflated, duration/survival, proportion/fractional, compositional, dyadic/network, and multilevel/panel). The skill includes multiple imputation via mice with Rubin's rules for combining estimates, Arellano--Bond GMM for dynamic panel models, Bayesian estimation via brms, latent class analysis, structural equation modeling, and sequence analysis. Each pipeline produces publication-ready tables, coefficient plots, and Methods/Results prose calibrated to journal conventions.

\textbf{Asset-driven writing with citation integrity.} \texttt{Scholar-write} uses a three-tier knowledge graph built from 127 published papers: 32 by the author, 8 collaborator articles, and 87 top-journal exemplars. Tier 1 provides an article knowledge base with pre-extracted stylistic annotations (opening hooks, gap sentences, contribution claims, voice registers, citation density norms, paragraph lengths) for all 127 papers, searchable by journal, method, or topic. Tier 2 provides a section-snippets library organized by nine rhetorical functions (hooks, gap statements, contribution claims, theory descriptions, methods leads, results leads, discussion openers, hedging, and quantitative sentence patterns). Tier 3 provides full PDF access for deep reads when the pre-built annotations are insufficient. Critically, before drafting begins, the skill builds a \textit{Verified Citation Pool}---a pre-approved list of references confirmed by searching the researcher's Zotero library. Only citations present in this pool may be inserted during drafting; the AI's training-data memory of citations is explicitly treated as unreliable. After drafting, a mandatory post-draft verification step cross-checks every inserted citation against the pool and converts any unverified reference to a \texttt{[CITATION NEEDED]} flag. The system generates original prose that mimics the \textit{rhetorical structure} of published papers while producing novel content---a distinction that separates structure mimicry from plagiarism. Each section undergoes a five-agent internal review panel---evaluating logic, rhetoric, journal fit, citation evidence, and accessibility---followed by a synthesizer that consolidates critiques with cross-agent agreement scoring and a reviser that implements changes, with user confirmation before advancing. Before drafting the Results section, the skill builds an \textit{Artifact Registry}---a numbered inventory of all pipeline-generated tables and figures from previous analysis phases. The Results section must reference every registered artifact with placement markers (e.g., \texttt{[Table N about here]}), and all tables and figures are consolidated at the manuscript end following journal conventions (ASR, AJS, and Demography place them on separate pages; Nature journals use Extended Data items).

\textbf{Peer review simulation.} \texttt{Scholar-respond} operates in five modes---simulate, respond, revise, resubmit, and cover-letter---supporting the full R\&R lifecycle. In simulate mode, it spawns three to seven reviewer agents in parallel---a methods specialist, a theory expert, a senior editor, and conditionally a computational methods reviewer, a demographics/population reviewer, a mixed-methods reviewer, and an ethics/responsible-AI reviewer---each calibrated to the target journal's priorities via a journal-specific emphasis table covering all twenty-two supported journals. The simulation produces a triage dashboard classifying concerns as MAJOR-FEASIBLE, MINOR-EASY, DISAGREE, or CONFLICT, with cross-reviewer overlaps flagged for priority attention. Within the orchestrator's Phase 10, the triage dashboard feeds a systematic section-by-section revision process: a \textit{Resolution Tracker} records the disposition of every categorized concern---RESOLVED (substantive revision made), REBUTTED (disagreement with rationale and minor concession), or DEFERRED (with justification)---and a verification gate requires all cross-reviewer items to be resolved before the manuscript advances to final assembly.

\textbf{Ethics and data safety.} \texttt{Scholar-safety} pre-scans data files for personally identifiable information using local pattern matching (no data enters the AI context), while \texttt{scholar-ethics} generates AI use disclosures, plagiarism audits, research integrity checks, and all journal-required declarations.

\textbf{Citation verification and management.} \texttt{Scholar-citation} operates in six modes: insert (add citations to uncited claims), audit (check for orphans, mismatches, and duplicates), convert-style (transform reference lists and in-text markers between seven citation styles including ASA, APA, Chicago, Nature, APSA, and Unified Linguistics), full-rebuild (regenerate the entire reference list from manuscript text), verify (5-tier systematic verification), and export (generate BibTeX \texttt{.bib} files). The verification mode implements a five-tier hierarchy: (1) local library search (Zotero, Mendeley, BibTeX, or EndNote), (2a) CrossRef API for DOI-based confirmation, (2b) Semantic Scholar API for preprints and working papers, (2c) OpenAlex API for open metadata covering 250M+ works, and (3) web search as a last resort. Each reference receives a status label---\texttt{VERIFIED-LOCAL}, \texttt{VERIFIED-CROSSREF}, \texttt{VERIFIED-S2}, \texttt{VERIFIED-OPENALEX}, \texttt{VERIFIED-WEB}, or \texttt{UNVERIFIED}---and unverified references are flagged for removal. This pipeline, combined with the Verified Citation Pool mechanism in \texttt{scholar-write}, implements a zero-tolerance policy for citation fabrication.

\textbf{Replication package construction.} \texttt{Scholar-replication} builds, documents, tests, and verifies a journal-ready replication package. It assembles scripts from the analysis pipeline into a structured directory with a master execution script, copies tables and figures from both the main analysis pipeline (\texttt{output/tables/}, \texttt{output/figures/}) and the EDA pipeline (\texttt{output/eda/tables/}, \texttt{output/eda/figures/}), and consumes the Artifact Registry produced by \texttt{scholar-write} as the authoritative table-and-figure numbering map. A format verification step confirms that each table has at least one renderable format (HTML, \TeX, or Word) and each figure is in a standard format (PDF or PNG). The skill generates an AEA-template README with nine sections, runs clean-run validation in an isolated environment comparing reproduced outputs against originals across both main and EDA directories, and performs a paper-to-code correspondence audit that consumes the Artifact Registry and manuscript placement markers (\texttt{[Table N about here]}) to map every table, figure, and in-text statistic to its producing script---flagging ORPHAN items (output exists but is not referenced in the manuscript) and MISSING items (referenced but no output file). Within the orchestrator, the replication package undergoes a final verification audit during assembly (Phase 11), ensuring paper-to-code correspondence is complete before submission. This distinguishes it from \texttt{scholar-open}, which produces open science \textit{declarations} (templates, checklists); \texttt{scholar-replication} \textit{actively builds} the package.

\textbf{Analytic action dispatch.} The orchestrator implements an Analytic Action Dispatch (AAD) protocol: when any phase produces an action item that requires running code---a robustness check, a power analysis, a marginal effects computation, a new figure---the system classifies the item as ANALYTIC, COMPUTATIONAL, or EXTERNAL. Analytic and computational items are dispatched to parallel task agents that write, execute, and return results automatically; only genuinely external tasks (e.g., ``collect new data,'' ``request IRB amendment'') remain as human action items. This protocol operates across all phases: power analysis in design, all EDA and regression tasks when data exists, reviewer-requested robustness checks in Phase 10, and any remaining analysis gaps during final assembly. The result is that in data-available mode, no analysis task is ever left as an author action item---the system executes all code and integrates results into the manuscript.

\textbf{Computational social science pipeline.} \texttt{Scholar-compute} provides ten specialized modules covering the major computational methods in contemporary social science: (1) text-as-data and NLP, including structural topic models, BERTopic with dynamic and hierarchical topics, text scaling via Wordfish and Wordscores, embedding regression via conText, LLM annotation with a four-risk framework \citep{lin2025navigating}, design-based supervised learning for bias-corrected downstream regression, multilingual NLP with cross-lingual embeddings via XLM-RoBERTa, and named entity recognition; (2) machine learning, including Double ML, causal forests, Bayesian regression with brms/Stan, and conformal prediction for distribution-free uncertainty quantification; (3) network analysis with ERGMs, SAOMs, relational event models, and graph neural networks via PyTorch Geometric; (4) agent-based modeling with Mesa 3.x, NetLogo, and LLM-powered agents; (5) computational reproducibility with renv, Docker, and Makefiles; (6) computer vision with DINOv2, CLIP, ViT, VideoMAE, and multimodal text-image fusion; (7) LLM-powered analysis workflows including structured extraction and computational grounded theory; (8) LLM synthetic data generation for silicon sampling, persona simulation, and survey simulation; (9) geospatial and spatial analysis with spatial lag, error, and SARAR models, Moran's I, and LISA maps; and (10) audio-as-data with Whisper transcription, speaker diarization, Essentia feature extraction, and acoustic feature analysis. Each module includes per-module verification subagents and produces publication-ready Methods and Results text.

\textbf{Continuous quality improvement.} \texttt{Scholar-auto-improve} provides a self-improving quality engine operating in four modes: \textit{observe} (monitor skill executions and log performance patterns), \textit{audit} (systematically evaluate skill outputs against quality checklists), \textit{improve} (propose and implement refinements to skill instructions, reference files, and quality gates), and \textit{evolve} (identify gaps in the skill ecosystem and generate new capabilities). This meta-skill enables the system to learn from each research engagement, progressively tightening quality gates and expanding coverage---a capacity absent from static tool pipelines.

\textbf{Qualitative and mixed methods.} \texttt{Scholar-qual} provides a complete qualitative research toolkit including inductive coding, grounded theory, thematic analysis, process tracing, and LLM-assisted coding with human-in-the-loop validation. The skill supports mixed-methods integration for sequential explanatory, convergent, and embedded designs, producing codebooks, audit trails, and inter-rater reliability statistics.

\textbf{Teaching materials.} \texttt{Scholar-teach} generates pedagogical content calibrated to course level (undergraduate, graduate, doctoral seminar): syllabi with session-by-session plans, lecture slide decks, problem sets with answer keys, exam questions with rubrics, and reading lists drawn from the researcher's Zotero library.

\textbf{Multi-author collaboration.} \texttt{Scholar-collaborate} manages multi-author workflows including CRediT (Contributor Roles Taxonomy) authorship statements, task assignment and tracking across co-authors, writing integration across independently drafted sections, and structured mentoring protocols for advisor--student collaborations.

\textbf{Dual execution modes.} The orchestrator operates in two modes depending on data availability. In \textit{data-available mode}, when the researcher provides existing data or the system can auto-fetch publicly available datasets (e.g., ACS via \texttt{tidycensus}, NHANES via \texttt{nhanesA}, World Bank via \texttt{WDI}), analysis tasks are dispatched to parallel task agents that write, execute, and return real results---no placeholders remain in the final manuscript. In \textit{code-template mode}, when data is not yet available, the system produces complete, runnable code labeled as templates, with table shells and placeholder cells, so that the full analytic pipeline is ready to execute once data arrives. This dual-mode architecture means the system never blocks on data availability but produces maximally complete output given the current state.

\textbf{Multi-format manuscript output.} The system produces manuscripts in four formats---Markdown, Word (.docx), \LaTeX, and PDF---via pandoc conversion, with all tables and figures appended at the manuscript end following journal conventions, ensuring compatibility with diverse journal submission systems and co-author workflows.

\subsection{What the System Is and Is Not}

Scholar-skill is not an autonomous research engine. It does not independently formulate research questions, make theoretical choices, or decide which findings are interesting. Every skill produces \textit{options}---candidate framings, alternative designs, draft prose---that the researcher must evaluate, select, and often revise. The system is best understood as a highly capable research assistant: one that can execute complex tasks across the pipeline but cannot replace the judgment that makes research meaningful.

This distinction is important because the same architecture that enables productive augmentation also enables uncritical delegation. A researcher who uses scholar-skill to generate five candidate research questions and selects the most theoretically interesting one is augmented. A researcher who accepts the first suggestion without evaluation is engaged in vibe researching in its most problematic form.

\section{A Cognitive Task Framework for AI-Augmented Research}

\subsection{Two Dimensions: Codifiability and Tacit Knowledge}

To move beyond generic claims about AI ``replacing'' or ``augmenting'' researchers, I propose a framework that classifies research tasks along two dimensions.

\textbf{Codifiability} refers to the extent to which a task can be decomposed into explicit, rule-following procedures \citep[cf.][]{cowan2000codification}. Literature search is highly codifiable: query databases, apply inclusion criteria, extract themes. Theory generation is weakly codifiable: it requires creative recombination of ideas in ways that resist algorithmic specification.

\textbf{Tacit knowledge requirement} refers to the extent to which successful task performance depends on knowledge that cannot be fully articulated---field politics, trust networks, timing intuition, editorial culture, a sense of what is ``live'' versus ``settled'' in a subfield \citep{collins2007rethinking}. \citeauthor{polanyi1966tacit}'s (\citeyear{polanyi1966tacit}) observation that ``we can know more than we can tell'' captures the structural limitation: AI systems operate on explicit, codified knowledge; they have no access to the tacit dimension.

\subsection{Four Task Types}

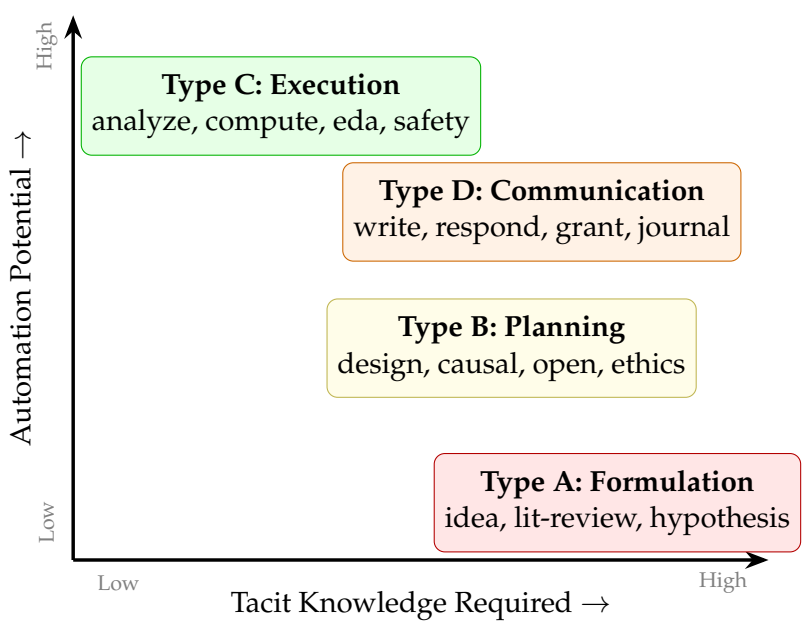
\begin{figure}[h]
\centering
\begin{tikzpicture}[scale=1.0]
  \draw[->, >=Stealth, line width=1.3pt] (0, 0) -- (9.2, 0);
  \draw[->, >=Stealth, line width=1.3pt] (0, 0) -- (0, 7.2);
  \node[font=\small] at (4.6, -0.6) {Tacit Knowledge Required $\rightarrow$};
  \node[font=\small, rotate=90] at (-0.7, 3.6) {Automation Potential $\rightarrow$};
  \node[font=\scriptsize, color=gray] at (0.6, -0.3) {Low};
  \node[font=\scriptsize, color=gray] at (8.6, -0.3) {High};
  \node[font=\scriptsize, color=gray, rotate=90] at (-0.35, 0.5) {Low};
  \node[font=\scriptsize, color=gray, rotate=90] at (-0.35, 6.8) {High};
  \node[rectangle, rounded corners=4pt, draw=green!70!black, fill=green!10,
        minimum width=3.2cm, minimum height=1.3cm, align=center, font=\small,
        anchor=west]
        at (0.1, 6.0) {\textbf{Type C: Execution}\\analyze, compute, eda, safety};
  \node[rectangle, rounded corners=4pt, draw=orange!80!black, fill=orange!10,
        minimum width=3.4cm, minimum height=1.3cm, align=center, font=\small]
        at (6.2, 4.6) {\textbf{Type D: Communication}\\write, respond, grant, journal};
  \node[rectangle, rounded corners=4pt, draw=yellow!70!black, fill=yellow!10,
        minimum width=3.2cm, minimum height=1.3cm, align=center, font=\small]
        at (5.8, 2.8) {\textbf{Type B: Planning}\\design, causal, open, ethics};
  \node[rectangle, rounded corners=4pt, draw=red!70!black, fill=red!10,
        minimum width=3.4cm, minimum height=1.3cm, align=center, font=\small,
        anchor=south]
        at (7.2, 0.1) {\textbf{Type A: Formulation}\\idea, lit-review, hypothesis};
\end{tikzpicture}
\caption{Four cognitive task types in the research pipeline, classified by automation potential and tacit knowledge requirement. Type C tasks (execution) are the most automatable. Type A tasks (formulation) are the least.}
\label{fig:taxonomy}
\end{figure}

Crossing these dimensions yields four task types, each with a different relationship to AI automation (Figure~\ref{fig:taxonomy}).

\textbf{Type C (Execution):} High automation potential, low tacit knowledge. Running regressions, producing descriptive statistics, generating visualizations, scanning data for sensitive information. AI agents are very good here---the tasks are codifiable and the outputs are verifiable.

\textbf{Type D (Communication):} Medium-high automation potential, medium tacit knowledge. Drafting prose, formatting for journals, responding to reviewers, writing grant proposals. AI can draft; the researcher must judge. This is the ``dangerous middle zone'' where AI output looks good but subtle errors require expert eyes \citep{dellacqua2023navigating}.

\textbf{Type B (Planning):} Medium automation potential, high tacit knowledge. Selecting identification strategies, constructing causal diagrams, designing open science protocols, making ethical judgments. The AI generates options; the researcher must own the choices.

\textbf{Type A (Formulation):} Low automation potential, very high tacit knowledge. Identifying new research questions, recognizing when existing frameworks are inadequate, importing framings from non-adjacent fields. The AI assists; the judgment is irreducibly human.

\subsection{The Delegation Boundary Is Cognitive, Not Sequential}

The framework reveals a critical insight: the boundary between what to delegate and what to protect is \textit{cognitive}, not \textit{sequential}. It does not fall between pipeline stages (``design is human, analysis is AI''). It cuts through every stage. At every stage, some tasks are codifiable and delegable; others require tacit judgment and are not. Table~\ref{tab:delegation} illustrates.

\begin{table}[h]
\centering
\caption{The Delegation Framework: What to Delegate, What to Protect}
\label{tab:delegation}
\small
\begin{tabular}{llll}
\toprule
\textbf{Task} & \textbf{Codifiable?} & \textbf{Tacit?} & \textbf{Delegate?} \\
\midrule
Literature synthesis & High & Low & Yes \\
Method selection \& code & High & Medium & Yes, with oversight \\
Analysis execution & High & Low & Yes \\
Prose drafting & Medium & Medium & Partially \\
RQ formulation & Low & High & AI assists only \\
Theory generation & Low & Very high & No \\
Field judgment & None & Extreme & Never \\
\bottomrule
\end{tabular}
\end{table}

The researcher who understands this framework is augmented---delegating execution while protecting judgment. The researcher who delegates everything is producing, at best, competent research with someone else's judgment. At worst, they are publishing errors they cannot detect.

\section{What AI Agents Can and Cannot Do}

\subsection{AI Excels at Speed and Coverage}

The speed advantage is substantial. A human researcher typically spends two to three weeks reading and synthesizing 100--200 papers on a topic, shaped by what they recall and biased toward familiar journals. \texttt{Scholar-lit-review-hypothesis} queries a 20,000-item Zotero library in seconds, produces thematic synthesis in under ten minutes, calibrates citation density to journal norms, and does so without fatigue or availability bias.

The productivity effects of generative AI on knowledge work are empirically documented. \citet{noy2023experimental} find that ChatGPT access led to large productivity gains in a randomized writing experiment, with effects concentrated among initially lower-performing workers. \citet{liang2024mapping} estimate that by early 2024, an estimated 17.5\% of computer science papers showed evidence of LLM-modified content. The trajectory is clear: AI-assisted research production is already widespread and growing.

The important caveat: coverage is not comprehension. The AI synthesizes what is in the database. It does not know what matters. It does not know which findings are credible, and it introduces systematic risks when used for tasks like annotation that require cultural and contextual judgment \citep{lin2025navigating}. It cannot sense the collective epistemic judgment of a field---what \citet{evans2011metaknowledge} call ``metaknowledge''---systematic knowledge about the structure and patterns of scientific knowledge itself, including the beliefs, preferences, and strategies embedded in research practices.

\subsection{AI Excels at Methodological Scaffolding}

A researcher trained in OLS and logistic regression can, with AI scaffolding, access working code for staggered difference-in-differences with \citet{callaway2021difference} estimators, exponential random graph models, causal forests with SHAP explanations, structural topic models, and zero-shot image classification with DINOv2 or CLIP. The AI provides working code, assumption statements, diagnostics, and write-up templates.

This scaffolding is genuinely valuable. It expands the set of methods a researcher can \textit{attempt}. But attempting a method and understanding it are different things. The researcher must still understand the assumptions well enough to defend them in peer review. A staggered DiD specification generated by \texttt{scholar-causal} is only as good as the researcher's judgment about whether the parallel trends assumption holds in their specific empirical context. That judgment is not in the code.

\subsection{AI Struggles with Theoretical Originality}

AI-generated theory looks like this: ``Drawing on Bourdieu's concept of field, I argue that X shapes Y through Z mechanism, moderated by Q.'' It is well-structured, correctly cited, and the mechanism is plausible. The problem: it is derived from existing frameworks. It recombines what is already there.

What AI cannot do: identify a genuinely new mechanism; recognize when existing frameworks are inadequate; import a framing from a non-adjacent field in a surprising way; make the conceptual leap that changes a subfield. Consider \citet{granovetter1973strength}'s ``strength of weak ties''---the argument was not derivable from the prior network literature. \citet{tilly1998durable}'s framework of durable inequality---in which exploitation and opportunity hoarding operate across categorical pairs---was not sitting in the literature waiting to be synthesized. These examples illustrate the irreducibly human zone that AI cannot currently enter.

\citet{si2024novel} find in a large-scale evaluation that LLM-generated research ideas received higher \textit{novelty} ratings than expert-generated ones but scored lower on \textit{feasibility} assessments. \citet{girotra2023ideas} find that LLMs generate more ideas at lower cost, and the best AI-generated ideas outperformed the best human ideas on average quality metrics---though with greater variance. These findings complicate a simple narrative: AI may generate surprising combinations, and can even produce top-ranked ideas, but the novelty measured in these studies is recombinative, not the deep theoretical innovation that characterizes paradigm-shifting work. The AI is an excellent theory \textit{articulator}; it is not a theory \textit{creator}. For most of what we publish, articulation is enough. For the papers that matter most, creation is what counts.

\subsection{AI Struggles with Tacit Field Knowledge}

The structural divide is between explicit knowledge (published literature, methods recipes, journal norms, data sources) and tacit knowledge (field politics, trust networks, timing intuition, editorial culture). The AI has the former in massive quantity. It has none of the latter.

Concretely: suppose you are writing about residential segregation. The AI can produce a competent research agenda, grounded in the literature, with sensible hypotheses. What it cannot tell you: whether the leading scholars in that subfield find your specific angle interesting or tired; whether this is a moment when segregation measurement is contested in ways that make your design politically fraught; whether a particular editor is sympathetic to structural versus behavioral arguments. That knowledge is embodied, relational, and accumulated through years of conference conversations and rejected papers \citep{collins2007rethinking}.

\section{Implications for the Social Sciences}

\subsection{The Augmentation Thesis and Its Fragile Condition}

In practice, AI augmentation looks like this: a researcher who previously mastered one or two methods can now, with scaffolding, access five or more; multiple datasets can be auto-fetched and explored simultaneously; the pipeline from idea to submission operates two to three times faster; reviewer responses take days instead of months.

Augmentation works \textit{when} the researcher remains the author, the judge, the intellectual owner. The AI is the research assistant, not the researcher. The human maintains the capacity for critical oversight. But this condition is fragile. \citet{raisch2021automation} formalize this as the ``automation--augmentation paradox'': automation and augmentation are interdependent, and overemphasizing either can fuel vicious cycles---including one in which successive delegation erodes the human capacity needed for effective augmentation. \citet{dellacqua2023navigating} demonstrate the danger in a field experiment with management consultants: GPT-4 increased performance on tasks inside the AI's capability frontier but \textit{decreased} performance on tasks outside it, because consultants over-relied on AI for tasks it could not do well. The ``jagged technological frontier'' they describe---where AI capability boundaries are unpredictable and non-obvious---applies directly to social science research.

This produces a \textit{verification gap}: if you did not participate in producing an output, you cannot reliably verify it. This is not a matter of trust in the AI; it is a matter of cognitive capacity. A researcher who has never manually coded interview transcripts cannot evaluate whether an AI's thematic analysis captures the right constructs. A researcher who has never hand-calculated a variance inflation factor cannot assess whether an AI's collinearity diagnostics are appropriate for the data structure. Verification requires the same expertise as production.

The question I would leave with the reader: if you never actually do the task---never run the DiD yourself, never draft the literature review by hand, never code the interviews---can you still evaluate whether the AI did it correctly? I believe the answer is no. And this is not a hypothetical future concern; it is a real pedagogical challenge right now.

\subsection{The Stratification Risk}

The productivity gains from AI tools are real. They are also unequally distributed. Four axes of stratification deserve attention.

\textbf{Cost.} Claude Pro is twenty dollars a month. API tokens for a full 16-phase pipeline run cost five to fifteen dollars. For researchers at resource-poor institutions, these are meaningful barriers.

\textbf{Language.} Training data skews English. Journal calibration targets English-language top journals. Non-English scholarship is disadvantaged by design.

\textbf{Technical skill.} Command-line setup, prompt engineering, and R/Python integration represent real barriers for researchers without computational training.

\textbf{Field.} The tool is calibrated to twenty-two journals spanning sociology, demography, political science, computational social science, and interdisciplinary science. While broader than earlier versions, area studies scholars, qualitative researchers, and Global South researchers whose work targets different venues get less value and sometimes actively incorrect guidance.

The ``AI productivity premium''---the widening gap between researchers with AI access and skills and those without---is itself a stratification phenomenon, one that social scientists are well positioned to study \citep{acemoglu2024simple, eloundou2024gpts, webb2020impact}. \citet{brynjolfsson2025canaries} document a 16\% relative employment decline among young workers (ages 22--25) in AI-automatable roles---the same dynamic threatens junior positions in academic research. It is also a call for design choices that prioritize access: open-weight models, documented prompts, and shared tools.

\subsection{The Pedagogical Crisis}

Traditional PhD training teaches students to run regressions, code interviews, write literature reviews, and respond to reviewers. These are execution skills. AI agents can now execute all of them. \citet{ferdman2025deskilling} documents that AI makes tasks seem cognitively easier, but workers are ceding problem-solving expertise to the system---what she terms ``capacity-hostile environments.'' In academic research, this dynamic is particularly dangerous because what \citet{hosseini2026benefits} describe as tasks ``closely connected to reasoning, communication, and imagination'' are precisely the ones AI automates first. This does not make these skills worthless---it changes what they are \textit{for}.

A student who understands difference-in-differences at a deep level---who has read Callaway and Sant'Anna, who understands the parallel trends assumption well enough to test it, who has manually worked through a staggered adoption case---can evaluate AI-generated DiD code. They can see when the parallel trends test is being misapplied. A student who only knows ``run the DiD package'' cannot.

The risk: if we train students to operate pipelines, and AI can operate pipelines, we are training them for a depreciating skill. The response is not to ban AI tools. It is to teach methods as foundations for \textit{evaluation}, not just \textit{production}. And it is to invest more heavily, not less, in deep theory training---because theory is precisely where AI is weakest and human scholars are irreplaceable \citep{bail2024generative}.

\section{Toward Responsible Vibe Researching}

I propose five principles for responsible use of AI agents in social science research.

\textbf{Disclose.} Report AI assistance in the methods section: what was used, when it was used, and which parts of the paper it contributed to. Example: ``Sections 3--5 were initially drafted with AI assistance using scholar-write; all content was reviewed and revised by the author.'' This is not a confession. It is a practice we need to normalize.

\textbf{Verify.} Review all AI-generated code, analysis, and prose before publishing. Run the code yourself. Check the citations against the 5-tier verification pipeline (local library, CrossRef, Semantic Scholar, OpenAlex, web search). Errors published under your name are yours. Scholar-skill includes verification sub-agents and a Verified Citation Pool that prevents citation fabrication during drafting, but these supplement---never replace---human oversight.

\textbf{Maintain skills.} Deliberately practice the tasks you delegate. Run regressions by hand sometimes. Write a section without AI sometimes. Not because AI assistance is cheating, but because oversight capacity depends on practiced judgment. The researcher who never does the work can no longer evaluate whether the AI did it correctly.

\textbf{Protect originality.} The research question and theoretical contribution must remain the researcher's own. That is not just a norm---it is where the intellectual value of the work lives. AI generates options; the researcher's theoretical imagination and field knowledge determine which options are worth pursuing.

\textbf{Design for access.} Use open models where feasible. Document prompts. Share tools. Do not let the AI productivity premium become another axis of academic stratification. The differential adoption of AI tools across institutions, languages, and fields is a social inequality that researchers can both study and mitigate through design choices.

\vspace{0.5em}

These five principles are complemented by four concrete practices for deliberate workflow design. First, \textit{map before you automate}: before adopting any AI tool, inventory the tasks in your research workflow and classify each along two dimensions---codifiability and tacit knowledge requirement. Automate high-codifiability, low-tacit tasks; protect low-codifiability, high-tacit tasks; for tasks in between, use AI as a drafting partner rather than a delegee. Second, \textit{maintain parallel competence}: even for tasks you delegate to AI, periodically perform them yourself---run a literature search manually alongside the AI's search, write a first draft before reading the AI's draft---to guard against the augmentation-to-dependency slide. Third, \textit{preserve apprenticeship pathways}: if you supervise students, identify which tasks serve primarily as training vehicles and protect those from automation, at least until the student has demonstrated mastery. The short-term efficiency loss is an investment in the student's long-term capacity and in the profession's ability to maintain standards of evaluation. Fourth, \textit{audit and document}: treat AI-assisted research the way we treat research involving human subjects---with transparency and accountability. Document which tasks were delegated to AI, what outputs were generated, and how those outputs were verified.

\section{Discussion and Conclusion}

This paper has made three arguments.

First, \textbf{AI agents are qualitatively different from prior tools}. They execute multi-step reasoning workflows---not just computation or text---across the entire research pipeline. The scholar-skill system demonstrates that a single plugin can now cover 26 distinct research tasks from idea formalization to journal submission, coordinated by an orchestrator across 18 phases with 53 quality gates and five hard stops. The system addresses a known weakness of LLM-based tools---citation fabrication---through a combination of proactive constraints (the Verified Citation Pool) and reactive verification (a 5-tier pipeline querying local libraries, CrossRef, Semantic Scholar, OpenAlex, and web search). The change in what a solo researcher can accomplish is real and significant.

Second, \textbf{the delegation boundary is cognitive, not sequential}. It cuts through every stage. At every point in the pipeline, some tasks are codifiable and delegable while others require tacit judgment and are not. The framework developed here---classifying tasks by codifiability and tacit knowledge requirement---provides a practical tool for researchers deciding what to delegate. The rule is simple: delegate codifiable execution; protect tacit judgment.

Third, \textbf{vibe researching is already here; the profession's normative response is not}. \citet{liang2024mapping} document the rapid spread of LLM-assisted writing in scientific papers. The systems are operational and improving. Disclosure norms, pedagogical reform, equitable design, and deeper theory training are the four urgent interventions. We are not waiting for a future that may arrive. We are in it.

Several limitations warrant acknowledgment. The case study is based on a single system (scholar-skill) developed by the author, which may not generalize to all AI research tools. The cognitive task framework, while grounded in the operational characteristics of the system, has not been empirically validated through user studies or controlled experiments. The four-type taxonomy simplifies a continuous space; some tasks may resist clean classification. Future work should empirically test the framework's predictions about delegation effectiveness, conduct user studies comparing augmented and unaugmented research workflows, and examine how AI tool adoption varies across disciplines, institutions, and career stages.

A useful analogy comes from aviation. The aviation industry did not achieve its safety record by telling pilots to ``just fly.'' It achieved it by building systems that preserved human judgment at critical decision points while automating everything else---years of training, rigorous examinations, standardized checklists, and an institutional infrastructure of air traffic control, maintenance protocols, and accident investigation. Research differs from aviation in that it demands novelty rather than standardization, but the principle holds: the most powerful tools require the most deliberate governance.

The wolf is at the door. The question is not whether to let it in---it is already in. The question is what role social scientists play, and what they protect. AI may transform social science research before social scientists have studied how AI transforms social science research. Our task is to use AI to amplify our reach while protecting our capacity for original inquiry---and to study AI itself as a social phenomenon shaping labor, inequality, and knowledge production.


\end{document}